\documentclass[paper]{article}
\usepackage{graphicx,xcolor}
\usepackage{float}
\usepackage[fleqn]{amsmath}
\usepackage{newtxtext}
\usepackage[varg]{newtxmath}

\usepackage{color}

\title{In-context
learning of closed form solution to simple linear regression task using 
transformer with linear self-attention}
\author{Katsuyuki Hagiwara}
\date{\normalsize Faculty of Education, Mie University\\
1577 Kurima-Machiya-cho, Tsu, 514-8507, Japan\\
hagi@edu.mie-u.ac.jp}

\usepackage{url}
\urldef{\mailsa}\path|hagi@edu.mie-u.ac.jp|

\def\q{{\boldsymbol{q}}}

\def\A{{\mathbf A}}
\def\O{{\mathbf O}}
\def\P{{\mathbf P}}
\def\Q{{\mathbf Q}}
\def\W{{\mathbf W}}
\def\H{{\mathbf H}}
\def\oH{\overline{\mathbf H}}

\def\LSA{{\rm LSA}}
\def\ox{{\overline{x}}}
\def\oy{{\overline{y}}}
\def\vxx{V}
\def\vxy{C}
\def\vTheta{{\boldsymbol{\Theta}}}

\def\ey{{\widehat{y}}}
\def\etheta{{\widehat{\theta}}}

\def\Win{{\mathbf W}_{\rm in}}
\def\Wout{{\mathbf W}_{\rm out}}
\def\bin{{\boldsymbol{b}_{\rm in}}}
\def\bout{b_{\rm out}}
\def\TF{{\rm TF}}
\def\D{{\mathbf D}}

\def\TFout{Q_{\rm out}}

\begin{document}
\maketitle
\begin{abstract}
In-context learning is a remarkable property of
transformers and has recently received a lot of interest. In many
studies of in-context learning, it has been shown that transformers are
capable of implementing solver for linear and non-linear regression
problems, in which the most of them implement gradient descent
algorithm. However, it is still unclear whether those implementations
have actually been acquired through training. In this paper, we
construct a transformer with linear self-attention, which in-context
learns the least squares estimate in a simple regression task. The point
here is that the closed form (analytical) solution is approximately
obtained by using layer normalization rather than an approximate
solution based on gradient descent algorithm. Then, we show an
experimental example, in which our implementation is mainly used in the
transformer trained with $\ell_1$ regularization when the target output
is the least squares estimate.

\end{abstract}

\section{Introduction}

In-context learning is a remarkable property of transformers, which form
the basis of large language models such as GPT-3 \cite{Brown2020}, and
it has been the focus of recent research. Through in-context learning,
given a prompt containing examples of a task and a new query input, 
the trained language model can generate the corresponding output for the new
query in a one-shot manner. It is natural to think that
transformers acquire, through training, the algorithms to solve tasks
via in-context learning. In this regard, it has been demonstrated that
transformers are capable of implementing various algorithms,
particularly in regression tasks
\cite{Garg2023,Oswald2022,Akyurek2022,Dai2023,Bai2023}.

\cite{Garg2023} empirically studied the in-context learning abilities of
transformers for various function classes in machine learning, including
the linear function class. In particular, for linear functions, a
trained transformer performs similarly to the least squares
solution. \cite{Oswald2022} provided an explicit construction of a
linear self-attention layer that implements a single step of the
gradient descent algorithm on the mean squared error loss. Additionally,
they empirically showed that several self-attention layers can
iteratively perform curvature correction, improving upon the plain
gradient descent algorithm. \cite{Akyurek2022} proved that a transformer
can implement a gradient descent algorithm and a closed form solution
for ridge regression.  \cite{Dai2023} also pointed out the
correspondence between the linear version of attention and the gradient
descent algorithm, claiming that transformers perform implicit
fine-tuning. They also empirically investigated the similarity between
in-context learning and explicit fine-tuning.  While the works of
\cite{Garg2023,Oswald2022,Akyurek2022,Mahankali2023,Dai2023} do not
consider the training phase, \cite{Zhang2023} investigated the learning
dynamics of a gradient flow in a simplified transformer architecture
when the training prompts consist of random instances of linear
regression datasets, concluding that transformers trained by a gradient
flow in-context learn a class of linear functions. More recently,
\cite{Bai2023} showed that transformers can implement a broad class of
standard machine learning algorithms in context, such as least squares,
ridge regression, and Lasso. In contrast to \cite{Akyurek2022},
\cite{Bai2023} precisely evaluated the prediction performance in terms
of network size, and shown a near-optimal predictive
power. \cite{Bai2023} also demonstrated the algorithm selection ability
of transformers, such as regularization selection according to
validation error for ridge regression. In these works, however, it still
remains unclear whether those algorithms are actually obtained through
the training process of transformers. In this paper, for a simple
regression task, we first construct a transformer with linear-self
attention, which implements an approximate closed form (analytical)
solution to the least squares estimate according to
\cite{Akyurek2022}. We then present a numerical example, in which this
implementation is mainly used in the transformer trained with the
$\ell_1$ regularization when the target is the least squares estimate.

Previous works \cite{Garg2023,Oswald2022,Akyurek2022,Dai2023,Bai2023}
have investigated regression problems, finding that the solver
implemented by a transformer is basically the gradient descent
algorithm. This is natural since the attention mechanism in a
transformer computes the product of the inputs, which is exactly what is
required for the gradient descent method. In a linear regression
problem, however, the least squares estimate is analytically obtained by
calculating the matrix inverse, which requires division. Consequently,
the gradient descent algorithm implicitly calculates this division by
repeatedly performing multiplication and addition. Among these works,
\cite{Akyurek2022} showed that a transformer can implement a close
form solution to a ridge regression problem, in which the division is
implemented using layer normalization. Our construction of the least
squares estimate for a simple regression problem is based on this
insight.

The transformer in our setting comprises a linear transformation of the
input, a stack of transformer blocks, a flatten layer, and an output
linear transformation, in which the transformer block consists of
multi-head linear self-attention blocks with layer normalization
followed by a skip connection.  Thus, the layer normalization is applied
to the sum of the output of the linear self-attention blocks and a skip
connection sum is applied to the layer normalized output. This
transformer receives the in-context samples and the query input
(prediction point) as a prompt, and outputs the least squares estimate
for the query input. Specifically, the number of layers, the number of
heads, and the model dimension in our construction are 2, 2, and 4
respectively, which is very small. The closed form solution to the least
squares estimate for a simple regression problem requires division by
the variance of the input data in the in-context samples. We provide a
specific construction to approximately calculate this closed form
solution under a natural input form, in which the layer normalization is
used for performing division according to \cite{Akyurek2022}. After
showing this construction, we present numerical experiments to show that
our implementation is actually used in the transformer trained with
$\ell_1$ regularization when the target output is the least squares estimate. In
other words, through training, the transformer mainly acquires the
calculation of the closed form solution rather than the steps of the
gradient descent algorithm.

This paper is organized as follows. Section 2 formulates our problem
setting. Section 3 presents the construction of a transformer that
calculates the least squares estimate based on in-context
samples. Section 4 provides a numerical example of training to
demonstrate that this construction is actually valid. Finally,
Section 5 concludes the paper and discusses future work.

\section{Problem setting}

\subsection{Notations}

In this paper, $\O_{I,J}$ is the $I\times J$ zero matrix. Note that we
use this notation also for vectors, where either $I$ or $J$ equals $1$.  For
an $I\times J$ matrix $\A$, $\A[i,j]$ is the $(i,j)$-entry of $\A$,
$\A[i,:]$ is the $i$-th row vector of $\A$ and $\A[:,j]$ is the $j$-th
column vector of $\A$ for $i=1,\ldots,I$ and $j=1,\ldots,J$. When $\A$
is an $I\times 1$ vector, $\A[i]$ denotes its $i$-th entry.

\subsection{Setting of in-context learning of transformer}

We next explain the transformer-based in-context learning of a simple
linear regression task. Let $M$ be the number of training data for the
transformer. Let $(x,y)$ be a pair of input-output variables in a
simple regression problem.  At each $m$ in $\{1,\ldots,M\}$,
a set of $N$ random samples of $(x,y)$, which is denoted by
$\{(x_{m,n},y_{m,n}):n=1,\ldots,N\}$, are generated.
We denote the $m$-th prompt (input to the transformer)
by $\P_m$, which is an $(N+1)\times 3$ matrix and whose $n$-th row is
\begin{align}
\label{eq:slr-prompt}
\P_m[n,:]=
\begin{bmatrix}
1 & x_{m,n} & y_{m,n}
\end{bmatrix}, 
\end{align}
where we define $y_{m,N+1}:=0$ and $x_{m,N+1}:=u_m$ that is a prediction
point; e.g., see \cite{Akyurek2022}. Thus, the input sequence length is $N+1$
and the number of in-context samples is $N$. We assume that, for each
$x_{m,n}$, $y_{m,n}$ is generated by
\begin{align}
y_{m,n}=\theta_{m,0}+\theta_{m,1}x_{m,n}+\varepsilon_{m,n}, 
\end{align}
where $\varepsilon_{m,1},\ldots,\varepsilon_{m,N}$, $m=1,\ldots,M$ are
i.i.d. additive noises from a probability distribution with mean $0$ and
variance $\sigma^2<\infty$. Our goal is to obtain the least squares
prediction at $x=u_m$ for the prompt $\P_m$.  Therefore, for each $m$, we need to
calculate the least squares solution using
\begin{align}
\label{eq:D_m}
\D_m=\{(x_{m,n},y_{m,n}):n=1,\ldots,N\},
\end{align}
which is the set of in-context samples. The important point is that the
regression lines can be different for each $m$. Thus, we may assume that
$(\theta_{m,0},\theta_{m,1})$ are sampled from a probability
distribution for each $m$. However, in the construction of the
transformer, we do not make any specific assumptions on the underlying
probability distribution of $(\theta_{m,0},\theta_{m,1})$. Although we
also do not make any specific assumptions on the underlying probability
distribution of $x_{m,n}$ and $\varepsilon_{m,n}$, we will make an
assumption on the in-context samples later.

\subsection{Training data for the transformer}

For the $m$-th training data $\D_m$ defined in (\ref{eq:D_m}), we define
\begin{align}
\ox_m&:=\frac{1}{N}\sum_{n=1}^Nx_{m,n}\\ 
\oy_m&:=\frac{1}{N}\sum_{n=1}^Ny_{m,n}\\ 
\label{eq:Vm}
\vxx_m&:=\frac{1}{N}\sum_{n=1}^N(x_{m,n}-\ox_m)^2\\ 
\label{eq:Cm}
\vxy_m&:=\frac{1}{N}\sum_{n=1}^N(x_{m,n}-\ox_m)(y_{m,n}-\oy_m).
\end{align}

For a simple regression problem, it is easy to see that the prediction at
$x=u_m$ using the least squares solution based on $\D_m$ is given by
\begin{align}
\label{eq:lse-estimate-at-u_m}
\ey_m(u_m):=\oy_m+\frac{\vxy_m}{\vxx_m}(u_m-\ox_m)
\end{align}
(see e.g., \cite{RD2002}).

Below, we construct the transformer that receives $\P_m$ as an input and 
outputs $\ey_m(u_m)$ for $m=1,\ldots,M$.
Thus, the transformer calculates the least squares estimate at
the prediction point $u_m$ using in-context samples $\D_m$ at each $m$.

\subsection{Linear self-attention}

We define a linear self-attention (LSA), which receives an $(N+1)\times
D$ matrix $\Q$ as an input and outputs $(N+1)\times
D$ matrix defined by
\begin{align}
\label{eq:LSA}
\LSA_{\vTheta}(\Q)&:=(\Q\W_3)(\Q\W_1^\top)^\top(\Q\W_2)\notag\\
&=(\Q\W_3)(\W_1\Q^\top\Q\W_2),
\end{align}
where $\vTheta=\{\W_1,\W_2,\W_3\}$ is an ordered set of parameters and
$\W_1$, $\W_2$ and $\W_3$ are $D\times D$ matrices. The operation using
this LSA is precisely discussed in \cite{Hagiwara2025}.

\subsection{Transformer structure}

\begin{figure}[ht]
\centering
\begin{minipage}{70mm}
\hspace{14mm}\centering
\includegraphics[width=40mm]{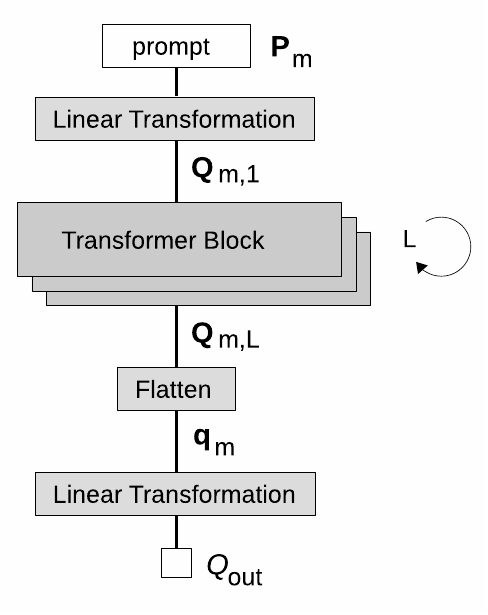}\\
(a) Main stream
\end{minipage}

\vspace{2mm}

\begin{minipage}{70mm}
\centering
\includegraphics[width=38mm]{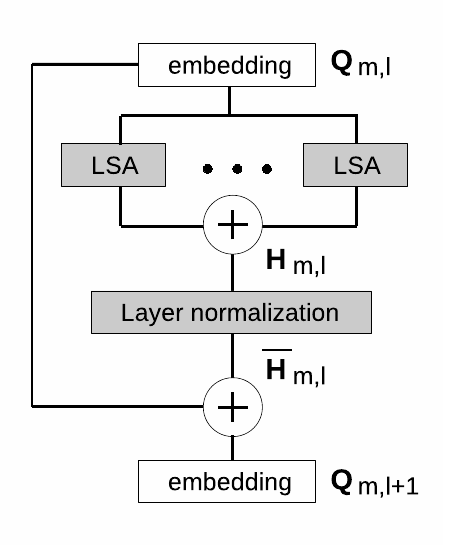}\\
(b) Transformer block
\end{minipage}

\caption{Structure of transformer}
\label{fig:structure}
\end{figure}

The transformer structure considered in this paper is illustrated in
Fig. \ref{fig:structure}, in which grey blocks are operation blocks. The
transformer here consists of input linear transformation, stacking
transformer blocks, flatten and output linear transformation
(Fig. \ref{fig:structure} (a)).  The transformer block consists of
multi-head LSA (MHLSA) block with layer normalization followed by a skip
connection (Fig. \ref{fig:structure} (b)).  It receives the $m$-th
prompt $\P_m$ defined in (\ref{eq:slr-prompt}) and, for simplicity,
outputs a scalar value whose target is the least squares estimate at a
prediction point $u_m$, which is $\ey_m(u_m)$ given by
(\ref{eq:lse-estimate-at-u_m}).

Let $\Win$ and $\bin$ be a $D\times 3$ input weight matrix and $D\times 1$
input bias vector respectively. 
The embedding of the prompt $\P_m$ 
by an affine transformation with $(\Win,\bin)$ is denoted by $\Q_{m,1}$ whose 
size is $(N+1)\times D$. More precisely, we define
\begin{align}
\label{eq:input-embedding}
\Q_{m,1}[n,:]^\top:=\Win\P_m[n,:]^\top+\bin
\end{align}
for $n=1,\ldots,N+1$.  The number of layers is denoted by $L$.  For the
$m$-th training data, the input to the $l$-th layer is denoted by
$\Q_{m,l}$, $l=1,\ldots,L$.  Thus, the above $\Q_{m,1}$ is the input to
the first layer.  

The parameter of the LSA block of the $k$-th head in the $l$-th layer is
denoted by $\vTheta_{l,k}=\{\W_{l,k,1},\W_{l,k,2},\W_{l,k,3}\}$. The
output of the MHLSA in the $l$-th layer for $m$-th training data is
\begin{align}
\label{eq:LSAoutput}
\H_{m,l,k}:=\LSA_{\vTheta_{l,k}}(\Q_{m,l})\\
\label{eq:MHLSAoutput}
\H_{m,l}:=\sum_{k=1}^{K}\H_{m,l,k},
\end{align}
where $K$ is the number of heads. 
The size of $\H_{m,l}$ is $(N+1)\times D$. The number of heads is common for all
layers. We employ the layer normalization along the model (embedding) dimension.
We define
\begin{align}
\label{eq:LN-mu}
\mu_{m,l,n}&:=\frac{1}{D}\sum_{d=1}^D\H_{m,l}[n,d]\\
\label{eq:LN-sigma2}
\sigma^2_{m,l,n}&:=\frac{1}{D}\sum_{d=1}^D
(\H_{m,l}[n,d]-\mu_{m,l,n})^2.
\end{align}
Then, the output of the layer normalization with these parameters is
denoted by $\oH_{m,l}$ whose $(n,d)$-entry is 
\begin{align}
\label{eq:after-LN}
\oH_{m,l}[n,d]:=(\H_{m,l}[n,d]-\mu_{m,l,n})/\sigma_{m,l,n}
\end{align}
for $d=1,\ldots,D$ at each $n$.  Finally, the output of the $l$-th layer
is given by
\begin{align}
\label{eq:after-skip-connection}
\Q_{m,l+1}=\Q_{m,l}+\oH_{m,l},
\end{align}
where the first term of the right-hand side comes from a skip
connection. Note that, in our model, the layer normalization is applied
before the skip connection(see e.g. \cite{Akyurek2022}).

$\Q_{m,L}$ is the output of the last layer. It is then flattened and
obtain $\q_m$ which is a $D(N+1)\times 1$ vector.  Let $\Wout$ and
$\bout$ be a $D(N+1)\times 1$ output weight matrix and output bias respectively.
We then obtain the transformer output by
\begin{align}
\label{eq:transformer-output}
\TFout(\P_m):=\Wout^\top\q_m+\bout
\end{align}
which is a scalar.

We refer to this transformer as $\TF(D,L,K)$, where $D$, $L$ and $K$ are
the model dimension, the number of layers and heads respectively. 

\section{A construction of transformer}

We here gives a construction of the transformer that calculates the
least squares estimate at the prediction point. To achieve this, we
require that $|u_m-\ox_m|/N$ and $|\oy_m|/N$ be negligible in the
calculation below. If we naturally assume that $|u_m-\ox_m|$ and
$|\oy_m|$ do not increase as $N$ increases then this is satisfied when
$N$ is sufficiently large. With this case in mind, we assume that $N$ is sufficiently
large so that $|u_m|/N$, $|\ox_m|/N$ and $|\oy_m|/N$ are negligible for
any $m$. We will address this assumption in a later section

We here fix the structure of the transformer, in which the model
dimension $D=4$, the number of layers $L=2$ and the number of heads $K=2$.
Therefore, the transformer is $\TF(4,2,2)$.  Now, the $m$-th prompt
to $\TF(4,2,2)$ is $\P_m$ defined in (\ref{eq:slr-prompt}).

\subsection{Input embedding}

Let $R$ be a real positive number.  We assume that $R$ can be chosen
sufficiently large below. As in \cite{Akyurek2022}, $R$ is important for
approximate implementation of the closed form solution.

In the input embedding, we set 
\begin{align}
\label{eq:Win}
\Win=\begin{bmatrix}
      1 & 0 & 0\\
0 & \frac{1}{RN} & 0\\
0 & 0 & \frac{1}{RN}\\
\frac{\sqrt{2}N}{N+1} & -\frac{1}{RN} & -\frac{1}{RN}
     \end{bmatrix} 
\end{align}
and $\bin=\O_{4,1}$.  By (\ref{eq:input-embedding}), the $n$-th row of
the input embedding is
\begin{align}
\label{eq:Q_m_1}
\Q_{m,1}[n,:]^\top
=
\begin{bmatrix}
1\\
\frac{1}{RN}x_{m,n}\\
\frac{1}{RN}y_{m,n}\\
\frac{\sqrt{2}N}{N+1}-\frac{1}{RN}x_{m,n}-\frac{1}{RN}y_{m,n}
\end{bmatrix},
\end{align}
which is the input to the transformer.

\subsection{Output of the first layer}

We here construct the first layer.

We consider the first head. We have
\begin{align}
\xi_{1,1,1}&:=\Q_{m,1}[:,1] ^\top\Q_{m,1}[:,1]
=N+1.
\end{align}
and
\begin{align}
\label{eq:xi_1_1_2}
\xi_{1,1,2}&:=\Q_{m,1}[:,1] ^\top\Q_{m,1}[:,2]\notag\\
&=\frac{1}{RN}(N\ox_m+u_m)\simeq\frac{1}{R}\ox_m,
\end{align}
where the last approximation holds 
by the assumption.
Since $y_{m,N+1}=0$, we have
\begin{align}
\xi_{1,1,3}&:=\Q_{m,1}[:,1] ^\top\Q_{m,1}[:,3]\notag\\
&=\frac{1}{RN}N\oy_m=\frac{1}{R}\oy_m.
\end{align}
By the same way as above, we also have
\begin{align}
\xi_{1,1,4}&:=\Q_{m,1}[:,1] ^\top\Q_{m,1}[:,4]\notag\\
&\simeq
\sqrt{2}N
-\frac{1}{R}\ox_m+
-\frac{1}{R}\oy_m.
\end{align}
Therefore, we have
\begin{align}
\Q_{m,1}^\top\Q_{m,1}=
\begin{bmatrix}
 \xi_{1,1,1} & \xi_{1,1,2} & \xi_{1,1,3} & \xi_{1,1,4}\\
 - & - & - & -\\
 - & - & - & -\\
 - & - & - & -\\
\end{bmatrix},
\end{align}
where the irrelevant elements are indicated by ``$-$''.
We set $\vTheta_{1,1}=\{\W_{1,1,1},\W_{1,1,2},\W_{1,1,3}\}$ in which
\begin{align}
\W_{1,1,1}&=\begin{bmatrix}
            1 & 0 & 0 & 0\\
            0 & 0 & 0 & 0\\
            0 & 0 & 0 & 0\\
            0 & 0 & 0 & 0\\
           \end{bmatrix}\\
\W_{1,1,2}&=\begin{bmatrix}
            \frac{N}{N+1} & 0 & 0 & 0\\
            0 & -\frac{1}{\sqrt{2}} & 0 & 0\\
            0 & 0 & -\frac{1}{\sqrt{2}} & 0\\
            0 & 0 & 0 & \frac{-1}{\sqrt{2}}\\
           \end{bmatrix}\\
\W_{1,1,3}&=\begin{bmatrix}
            1 & 0 & 0 & 0\\
            0 & 0 & 0 & 0\\
            0 & 0 & 0 & 0\\
            0 & 0 & 0 & 0\\
           \end{bmatrix}.
\end{align}
Then, by (\ref{eq:LSA}) and (\ref{eq:LSAoutput}), the $n$-th row of the
output of the first head is
\begin{align}
\H_{m,1,1}[n,:]^\top
=\begin{bmatrix}
N\\  
-\frac{1}{\sqrt{2}R}\ox_m\\
-\frac{1}{\sqrt{2}R}\oy_m\\
-N+\frac{1}{\sqrt{2}R}\ox_m
+\frac{1}{\sqrt{2}R}\oy_m
 \end{bmatrix}.
\end{align}

On the other hand, for the second head, we set 
$\vTheta_{1,2}=\{\W_{1,2,1},\W_{1,2,2},\W_{1,2,3}\}$ in which
\begin{align}
\W_{1,2,1}=\W_{1,2,2}=\W_{1,2,3}=\O_{d,d}.
\end{align}
Then, by (\ref{eq:LSA}) and (\ref{eq:LSAoutput}), the $n$-th row of the
output of the second head is
\begin{align}
\H_{m,1,2}[n,:]=\O_{1,4}. 
\end{align}

Therefore, by (\ref{eq:MHLSAoutput}), the $n$-th row of the output of
the MHLSA in the first layer is
\begin{align}
\label{eq:const-MHLSAoutput1}
\H_{m,1}[n,:]^\top&=\H_{m,1,1}[n,:]^\top\notag\\
&=
\begin{bmatrix}
N\\
-\frac{1}{\sqrt{2}R}\ox_m\\
-\frac{1}{\sqrt{2}R}\oy_m\\
-N+\frac{1}{\sqrt{2}R}\ox_m
+\frac{1}{\sqrt{2}R}\oy_m
\end{bmatrix}
\end{align}

Now, we consider the layer normalization. 
By (\ref{eq:LN-mu}), (\ref{eq:LN-sigma2}) and 
(\ref{eq:const-MHLSAoutput1}), for $\H_{m,1}[n,:]$, we have
\begin{align}
\label{eq:mu_m_1_n}
\mu_{m,1,n}&=0\\
\label{eq:sigma2_m_1_n}
\sigma^2_{m,1,n}&=\frac{1}{4}\left(
N^2+O\left(\frac{1}{R^2}\right)+N^2+O\left(\frac{N}{R}\right)\right)
\simeq \frac{N^2}{2},
\end{align}
since we can set a sufficiently large value for $R$; e.g.,  $R=N^p$,
where $p$ is a sufficiently large positive integer. 

By (\ref{eq:after-LN}),
(\ref{eq:const-MHLSAoutput1}), 
(\ref{eq:mu_m_1_n}) and 
(\ref{eq:sigma2_m_1_n}), the $n$-th row of the output of the layer
normalization is
\begin{align}
\label{eq:oH_m_1}
\oH_{m,1}[n,:]^\top=
\begin{bmatrix}
 \sqrt{2}\\
-\frac{1}{RN}\ox_m\\
-\frac{1}{RN}\oy_m\\
-\sqrt{2}
+\frac{1}{RN}\ox_{m,n}
+\frac{1}{RN}\oy_{m,n}
\end{bmatrix}. 
\end{align}

Then, by (\ref{eq:after-skip-connection}), (\ref{eq:Q_m_1}) and
(\ref{eq:oH_m_1}), the $n$-th row of the output of the first 
layer is
\begin{align}
\label{eq:Q_m_2}
\Q_{m,2}[n,:]^\top=
\begin{bmatrix}
1+ \sqrt{2}\\
\frac{1}{RN}\Delta x_{m,n}\\
\frac{1}{RN}\Delta y_{m,n}\\
-\frac{\sqrt{2}}{N+1}
\end{bmatrix},
\end{align}
where
\begin{align}
\label{eq:Delta_x_m_n}
\Delta x_{m,n}&=x_{m,n}-\ox_m\\ 
\label{eq:Delta_y_m_n}
\Delta y_{m,n}&=y_{m,n}-\oy_m.
\end{align}
and the $O(1/(RN))$ terms in $\Q_{m,2}[n,4]$ is omitted since 
$R$ is sufficiently large and, indeed, they do not affect
the final outcome in our construction.

\subsection{Output of the second layer}

We next construct the second layer.

By (\ref{eq:Q_m_2}), we have
\begin{align}
\xi_{2,2,2}&:=Q_{m,2}[:,2]^\top Q_{m,2}[:,2]\notag\\
&=\frac{1}{R^2N^2}\{N\vxx_m+(u_m-\ox_m)^2\}\notag\\
&\simeq\frac{1}{R^2N}\vxx_m\\
\xi_{2,2,3}&:=Q_{m,2}[:,2]^\top Q_{m,2}[:,3]\notag\\
&=\frac{1}{R^2N^2}\{N\vxy_m+(u_m-\ox_m)(0-\oy_m)\}\notag\\
&\simeq\frac{1}{R^2N}\vxy_m
\end{align}
under the assumption and, thus, we have
\begin{align}
\label{eq:Q_m_2TQ_m_2}
\Q_{m,2}^\top\Q_{m,2}=\begin{bmatrix}
- & - & - & -\\
- & \xi_{2,2,2} & \xi_{2,2,3} & - \\
- & - & - & - &\\
- & - & - & - &\\
\end{bmatrix}.
\end{align}

We consider the first head. We 
set $\vTheta_{2,1}=\{\W_{2,1,1},\W_{2,1,2},\W_{2,1,3}\}$, in which
\begin{align}
\W_{2,1,1}&=
\begin{bmatrix}
0 & 1 & 0 & 0\\
0 & 0 & 0 & 0\\
0 & 0 & 0 & 0\\
0 & 0 & 0 & 0\\
\end{bmatrix}\\
\W_{2,1,2}&=
\begin{bmatrix}
0 & 0 & 0 & 0\\
0 & 0 & 0 & 0\\
0 & R^2N/\sqrt{2} & -R^2N/\sqrt{2} & 0\\
0 & 0 & 0 & 0\\
\end{bmatrix}\\
\W_{2,1,3}&=
\begin{bmatrix}
0 & 0 & 0 & 0\\
1 & 0 & 0 & 0\\
0 & 0 & 0 & 0\\
0 & 0 & 0 & 0\\
\end{bmatrix}.
\end{align}
Then, by (\ref{eq:LSA}), (\ref{eq:LSAoutput}) and
(\ref{eq:Q_m_2TQ_m_2}), the $n$-th row of the output of the first head
is
\begin{align}
\H_{m,2,1}[n,:]^\top
=\begin{bmatrix}
0\\
\frac{1}{\sqrt{2}RN}\vxy_m\Delta x_{m,n}\\
-\frac{1}{\sqrt{2}RN}\vxy_m\Delta x_{m,n}\\
0\\
 \end{bmatrix}.
\end{align}

On the other hand, we consider the second head.
We set $\vTheta_{2,2}=\{\W_{2,2,1},\W_{2,2,2},\W_{2,2,3}\}$ in which
\begin{align}
\W_{2,2,1}&=
\begin{bmatrix}
0 & 1 & 0 & 0\\
0 & 0 & 0 & 0\\
0 & 0 & 0 & 0\\
0 & 0 & 0 & 0\\
\end{bmatrix}\\
\W_{2,2,2}&=
\begin{bmatrix}
0 & 0 & 0 & 0\\
R^2N & 0 & 0 & -R^2N\\
0 & 0 & 0 & 0\\
0 & 0 & 0 & 0\\
\end{bmatrix}\\
\W_{2,2,3}&=
\begin{bmatrix}
1/(1+ \sqrt{2}) & 0 & 0 & 0\\
0 & 0 & 0 & 0\\
0 & 0 & 0 & 0\\
\end{bmatrix}.
\end{align}
Then, by (\ref{eq:LSA}), (\ref{eq:LSAoutput}) and
(\ref{eq:Q_m_2TQ_m_2}), the $n$-th row of the output of the second head
is
\begin{align}
\H_{m,2,2}[n,:]^\top
=\begin{bmatrix}
\vxx_m\\
0\\
0\\
-\vxx_m\\
 \end{bmatrix}.
\end{align}

Therefore, by (\ref{eq:MHLSAoutput}), the $n$-th row of the output of
the MHLSA in the second layer is
\begin{align}
\label{eq:const-MHLSAoutput2}
\H_{m,2}[n,:]^\top
=\begin{bmatrix}
\vxx_m\\
\frac{1}{\sqrt{2}RN}\vxy_m\Delta x_{m,n}\\
-\frac{1}{\sqrt{2}RN}\vxy_m\Delta x_{m,n}\\
-\vxx_m\\
 \end{bmatrix}.
\end{align}

Now, we consider the layer normalization. 
By (\ref{eq:LN-mu}), (\ref{eq:LN-sigma2}) and 
(\ref{eq:const-MHLSAoutput2}), for $\H_{m,2}[n,:]$, we have
\begin{align}
\mu_{m,2,n}&=0\\
\label{eq:sigma2_m_2_n}
\sigma^2_{m,2,n}&=\frac{1}{4}\left(
2\vxx_m^2+O\left(\frac{1}{R^2N^2}\right)\right)
\simeq \frac{1}{2}\vxx_m^2
\end{align}
since $R$ is sufficiently large.
By (\ref{eq:after-LN}), the $n$-th row of the output of the layer normalization is
\begin{align}
\label{eq:oH_m_2}
\oH_{m,2}[n,:]^\top=
\begin{bmatrix}
\sqrt{2}\\
\frac{1}{RN}\frac{\vxy_m}{\vxx_m}\Delta x_{m,n}\\
-\frac{1}{RN}\frac{\vxy_m}{\vxx_m}\Delta x_{m,n}\\
-\sqrt{2}\\
\end{bmatrix}. 
\end{align}
The division performed by the layer normalization is actually used for
this purpose in \cite{Akyurek2022}.

Then, by (\ref{eq:after-skip-connection}), (\ref{eq:Q_m_2}) and
(\ref{eq:oH_m_2}), the $n$-th row of the output of the second layer is
\begin{align}
\label{eq:Q_m_3}
\Q_{m,3}[n,:]^\top=
\begin{bmatrix}
1+ 2\sqrt{2}\\
\frac{1}{RN}\Delta x_{m,n}+
\frac{1}{RN}\frac{\vxy_m}{\vxx_m}\Delta x_{m,n}\\
\frac{1}{RN}\Delta y_{m,n}
-\frac{1}{RN}\frac{\vxy_m}{\vxx_m}\Delta x_{m,n}\\
-\sqrt{2}(N+2)/(N+1)\\
\end{bmatrix}.
\end{align}

\subsection{Model output}

Since $x_{m,N+1}=u_m$ and $y_{m,N+1}=0$, we have
\begin{align}
\label{eq:Q_m_3-N+1_3}
\Q_{m,3}[N+1,3]=-\frac{1}{RN}
\left(\oy_m+\frac{\vxy_m}{\vxx_m}(u_m-\ox_m)\right) 
\end{align}
by (\ref{eq:Q_m_3}).  
Now, $\Q_{m,3}$ is flattened into $\q_m$.
Then, we have $\q_m[D(N+1)-1]=\Q_{m,3}[N+1,3]$.
Hence, if we set
\begin{align}
\label{eq:Wout}
\Wout[i]=\begin{cases}
          -RN & i=D(N+1)-1\\
0 & {\rm otherwise}
         \end{cases}
\end{align}
then the transformer output for the $m$-th prompt is, approximately, 
\begin{align}
\label{eq:final-tf-output}
\TFout(\P_m)=\oy_m+\frac{\vxy_m}{\vxx_m}(u_m-\ox_m)
\end{align}
by (\ref{eq:transformer-output}) and (\ref{eq:Q_m_3-N+1_3}).  This is
consistent with the least squares estimate at $u_m$ in
(\ref{eq:lse-estimate-at-u_m}).

\subsection{Discussion}
\label{sec:discussion}

In our construction, we need to choose a large value for $N$ that is the
number of in-context samples. This is because, for example as in
(\ref{eq:xi_1_1_2}), we need to eliminate the effect of the prediction
point $(u_m,0)$ on the calculation of the mean of the in-context input
samples. In other words, it is necessary to treat the in-context samples
and prediction point separately here.  This is not possible since the
attention mechanism calculates the inner product of the two sequences,
which implies that information on the sequence dependent property
disappears; e.g., see (\ref{eq:LSA}). In large language models, this fact
is generally important since we may need masking and/or positional
encoding to control the sequence dependent property that cannot be
controlled by the attention mechanism.

Note that if we appropriately design the prompt (input to the
transformer) then we may not need to worry about this approximation
problem.  For example, in the prompt, the in-context samples and
prediction point are embedded separately in the different sequences. The
input employed in our setting is just a natural one that is also
employed in \cite{Akyurek2022}.  In other words, the prompt (input)
design is possible to affect the algorithm that is obtained by
training. This is also pointed out in \cite{Hagiwara2025}.  Therefore, a
large $N$ assumption may not be essential for performing division
using the layer normalization. The construction under the different
prompt design is left as a future work.

On the other hand, a large value for $R$ is significantly required for
performing division by layer normalization.  Note that this enforces
the absolute values of weights to be either close to zero or extremely
large.

Our construction may be relatively compact.
Unfortunately, the representation of LSA is not unique in the sense
that, as shown in \ref{sec:appendix-nonuniqueness}, it produces the same output
for different inputs.  Therefore, it might be possible to create the
same output using a more verbose expression, in which the most weight
values are not zeros.

\section{Numerical experiment}

We here show several numerical experiments to demonstrate that our
construction in the previous section is actually used in the trained transformer.

\subsection{Setting of experiment}

The transformer defined in this paper is coded by PyTorch, in which we
omit the biases in the input and output linear transformation using {\tt
torch.nn.Linear}.  For the training data $\D_m$ defined by
(\ref{eq:D_m}), we set $\sigma^2=0.04$ for the noise variance and, at
each $m$, the inputs including prediction points are randomly drawn from
a standard normal distribution $N(0,1)$. And, we generate $\theta_{m,0}$
and $\theta_{m,1}$ according to $N(0,1)$ independently at each $m$.  The
length of the input sequence is $N+1=51$ (the number of in-context
samples is $N=50$). The target output (teacher) for the transformer is
the least squares estimate at the prediction point.  This is a more
direct instruction than the output data at the prediction point. The transformer
is $\TF(4,2,2)$.  For the training, the number of training data is
$M=5000$, the number of validation data is $1000$, the batch size is
$500$, the learning rate is $0.005$, the maximum number of training
epochs is $20000$.  We apply the layer normalization without an
element-wise affine setting in {\tt torch.nn.LayerNorm}.  As mentioned
in Sec. \ref{sec:discussion}, there is a lack of uniqueness in the LSA
representation. Therefore, we apply the $\ell_1$ regularization to the
input and output weights, by which non-contributing weights are set to
zero and important weights may be highlighted. This may make one expect
a relatively compact expression as obtained in our construction.  We
employ the transformer that minimizes the validation error calculated
every $100$ epochs.

\subsection{Verifying the use of layer normalization}
\label{sec:prediction_correlation}

We execute $10$ different training runs under the above setting, in
which both of data and initialization differ for each run.  We generate
$1000$ new data of pairs of the prompt and the least squares estimate,
in which the prompts are i.i.d. samples from the same distribution as
the prompt in the training data. Then, for each prompt, we obtain the
output of the trained transformer and can calculate the squared test
error between the output and the least squares estimate at the
prediction point. Thus, in each run, we can obtain the average test error
for $1000$ new data.

Simultaneously, for each $m$-th prompt in the new data, we can calculate
$\H_{m,2}$ defined in (\ref{eq:MHLSAoutput}).  The variance along the
embedding used in the layer normalization is the variance of
$\H_{m,2}[n,:]$ for the $n$-th sequence, which is $\sigma_{m,2,n}^2$
defined in (\ref{eq:LN-sigma2}). Also, we can calculate $\vxx_m$ defined
in (\ref{eq:Vm}) as the variance of in-context input samples in the new
data.  In our construction, the division by $\vxx_m$ in
(\ref{eq:lse-estimate-at-u_m}) is executed using the division by the
standard deviation in the layer normalization in (\ref{eq:oH_m_2}).
To see this, we calculate the correlation coefficient between $1/\sigma_{m,2,51}$,
$m=1,\ldots,1000$ and $1/\vxx_m$ , $m=1,\ldots,1000$ in each run.

The results for $10$ runs are summarized in Table
\ref{tbl:error-correlation}. We can see that the correlation coefficient
is larger than $0.9$ in the first and ninth runs. This implies that the
division by $V_m$ is approximately calculated using the layer
normalization for these two runs. Since, as seen in our construction,
there exist approximation errors in doing this, these correlation
coefficients may be satisfactory. Additionally, the trained transformer
in the first run shows a better predictive performance in terms of the
average test error.  Then, we further analyze the trained transformer in
the first run below. Note that the trained transformer in the fourth run
shows the best predictive performance while the correlation is low. This
indicates that an alternative way of the construction exists. We discuss
this point in the conclusion.

\begin{table}[h]
\begin{center}
\caption{Average squared test error and correlation between reciprocals
of standard deviation in layer normalization and variance of in-context input samples}
\label{tbl:error-correlation}
\begin{tabular}{|c|c|c|}\hline
No. & Average test error & Correlation\\\hline
1 & 0.00072  &  0.90608 \\\hline
2 & 0.00139  &  -0.00510 \\\hline
3 & 0.00088  &  0.60064 \\\hline
4 & 0.00063  &  0.50725 \\\hline
5 & 0.00267  &  0.20170 \\\hline
6 & 0.00165  &  0.55747 \\\hline
7 & 0.00092  &  0.53876 \\\hline
8 & 0.00074  &  0.10411 \\\hline
9 & 0.00105  &  0.92632 \\\hline
10 & 0.00131  &  0.73416 \\\hline
\end{tabular}
\end{center}
\end{table}

\subsection{Input and output weights}

\begin{figure}[ht]
\centering
\begin{minipage}{80mm}
\centering
\includegraphics[width=65mm]{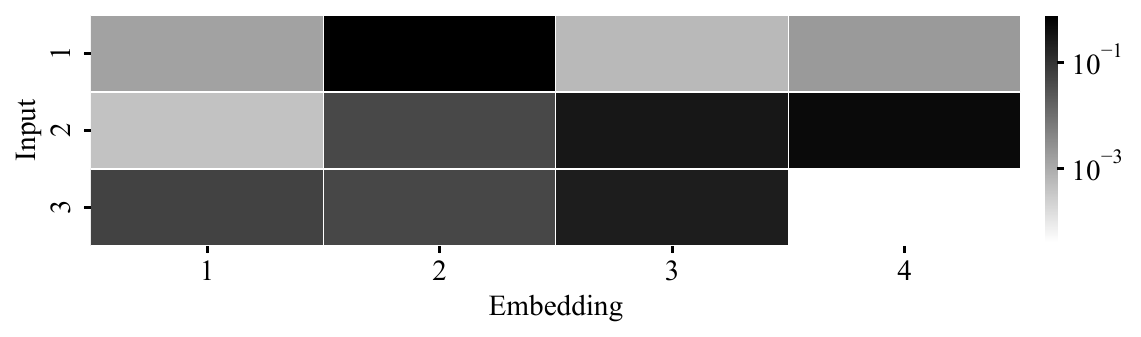}\\
(a) Input weights
\end{minipage}

\vspace{2mm}

\begin{minipage}{80mm}
\centering
\includegraphics[width=80mm]{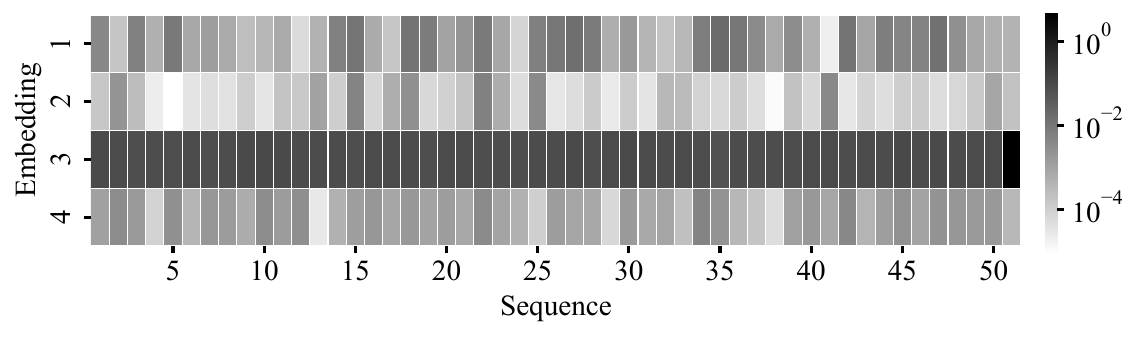}\\
(b) Output weights
\end{minipage}

  \caption{Absolute weight values}
  \label{fig:weight_value}
\end{figure}

In Fig. \ref{fig:weight_value} (a) and (b), for the first run, we show
the absolute values of the input and output weights which are denoted by
$\Win$ and $\Wout$ in this paper. The weight values are shown on a
logarithmic scale.  $\Win$ is a $3\times 4$ matrix. The row and column
are the indices of inputs and embeddings respectively.  For the input
weights, the indices of inputs $(1,2,3)$ correspond to the variables
$(1,x,y)$ in the prompt to the transformer; i.e., a constant term, input
and output in a simple regression problem.  Although $\Wout$ is a
$204$-dimensional vector, we here convert it into a $4\times 51$ matrix
whose row and column correspond to the indices of embeddings and
sequences respectively.  We refer to the entry in row $i$ and column $j$
of these weight matrices as the $(i, j)$ weight value. Especially, the
output weight is the weight for the output of the cell in the final
layer; i.e., $(i,j)$ weight is for the cell that is located in the $j$-th
sequence of the $i$-th embedding. We refer to the cell connected to the
$(i,j)$ weight as the $(i,j)$ cell.

In Fig. \ref{fig:weight_value} (a), for the first embedding that
corresponds to the first column in the figure, the $(3,1)$ value is
larger than $(1,1)$ and $(2,1)$ values, which implies that $y$ is
dominant in this embedding. For the second embedding, the $(1,2)$ value
is relatively larger than the other values, which implies that the
constant term is dominant in this embedding.  For the third embedding,
the $(2,3)$ and $(3,3)$ values are larger than the $(1,3)$ value, which
implies that $(x,y)$ are dominant in this embedding.  Lastly, for the
fourth embedding, the $(2,4)$ value is larger than the other values,
which implies that $x$ is dominant in this embedding.  Although the
non-dominant weights are not exactly zero, this result may support that
the inputs are extracted in the embedding separately for the further
calculation. Note that this representation may vary depending on the run
since LSA is possible to give the same output for different inputs;
e.g., see \ref{sec:appendix-nonuniqueness}. 

In Fig. \ref{fig:weight_value} (b), we can see that the weight values of
the third embedding cells are relatively large and, especially, the
$(3,51)$ value is extremely large compared to the other values.
Therefore, the $(3,51)$ cell is the dominant factor for the transformer
output. Note that the contribution of the third embedding cells are
relatively large compared to the other embeddings. The effect of those
cells will be mentioned in the conclusion.

\subsection{Regression analysis}

We here check that, in the first run, the output of the $(3,51)$ cell in
the second layer has the form of (\ref{eq:lse-estimate-at-u_m}) by using
a regression analysis.  We generate $1000$ new data for the prompt to
the trained transformer, which are i.i.d. samples from the same
distribution as the training data. Then, for each $m$-th new prompt, we
can obtain $\Q_{m,3}$ in (\ref{eq:Q_m_3}) and focus on
$z_m:=\Q_{m,3}[51,3]$, which is the output of the $(3,51)$ cell in the
second layer. We then fit $z_m$ by
\begin{align}
f_m:=\alpha_0+\alpha_1(u_m-\ox_m)
+\alpha_2\oy_m
+\alpha_3\frac{C_m}{V_m}(u_m-\ox_m),
\end{align}
where $(\alpha_0,\alpha_1,\alpha_2,\alpha_3)$ are regression
coefficients. 

\begin{table}[h]
\begin{center}
\caption{Results of regression analysis}
\label{tbl:estimated-coef}
\begin{tabular}{|c|c|c|c|}\hline
$\alpha_0$ & $\alpha_1$ & $\alpha_2$ & $\alpha_3$\\\hline
0.00250  & -0.00159 & -0.2436  & -0.22498\\
 & (-0.00453) & (-0.72452) & (-0.69403)\\\hline
\end{tabular}
\end{center}
\end{table}

The coefficient estimates are summarized in Table
\ref{tbl:estimated-coef}, in which we show the standardized regression
coefficients in the brackets. In Table \ref{tbl:estimated-coef}, the
estimates of $\alpha_0$ and $\alpha_1$ are small and those of $\alpha_2$
and $\alpha_3$ are large. Here, the correlation between $z_m$ and $f_m$
was $0.99479$. Thus, $z_m$ is found to be the form of the last two
terms in $f_m$.  It is the same form of the least squares solution
in (\ref{eq:lse-estimate-at-u_m}).  And, the correlation between
$z_m$ and the least squares solution was $-0.99396$, which implies that
the output of the $(3,51)$ cell is almost consistent with the least
quares solution except the sign.  The sign can be changed by adjusting
the corresponding output weight as in our construction.

\subsection{Prediction}

We finally show the fitting curve of the trained transformer in the
first run to visually confirm that our conclusion is valid. We generate
$100$ new data for the prompts to the trained transformer, which are
$\{(x_1,y_1),\ldots,(x_n,y_{50}),(u_k,0)\}$ for $k=1,\ldots,100$, where
$(x_i,y_i)$ is generated from the same distribution as the training data
and $u_k$ is the equidistant points in $[-3,3]$.  In other words, we
obtain the transformer output at $100$ different prediction points under
the same in-context samples.  Therefore, we can plot a fitting curve for
the in-context samples.

\begin{figure}[ht]
  \centering
\begin{minipage}{50mm}
\centering
\includegraphics[width=41mm]{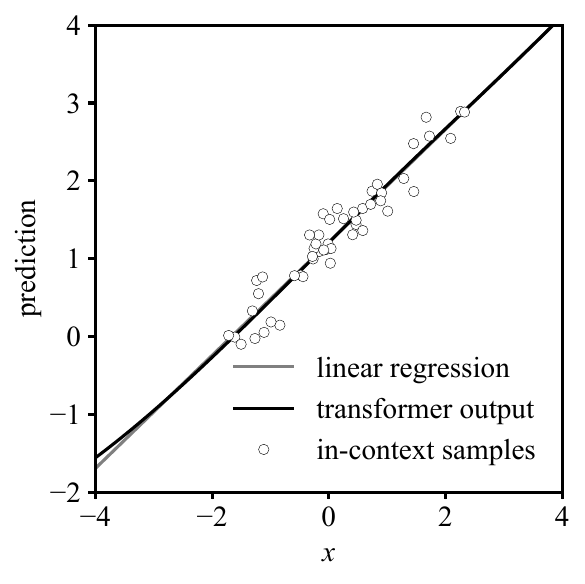}\\
(a) Transformer output
\end{minipage}
\begin{minipage}{50mm}
\centering
\includegraphics[width=41mm]{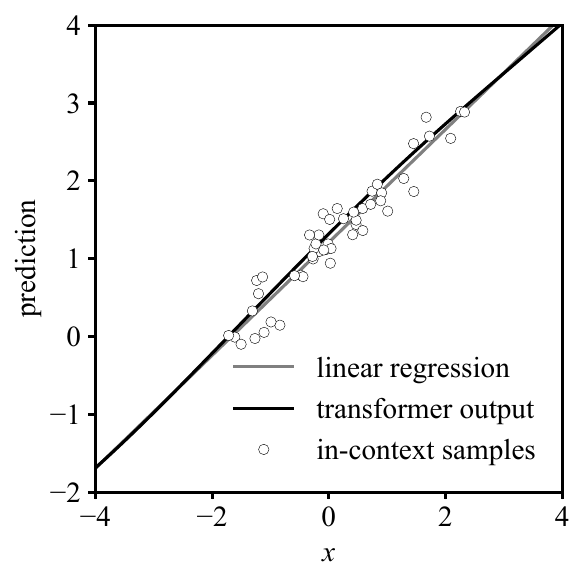}\\
(b) Output through $(3,51)$ cell
\end{minipage}
 \caption{Fitting curve}
  \label{fig:fitting_example}
\end{figure}

In Fig. \ref{fig:fitting_example} (a), we show the in-context samples
(open circle) and the transformer output (black solid line). In this
figure, we also show the least squares estimate of a simple regression
model (gray solid line), which is obtained for the in-context samples,
$(x_1,y_1),\ldots,(x_n,y_{50})$. We can see that the transformer output
is well consistent with the least squares estimate of a simple
regression model. In Fig. \ref{fig:fitting_example} (b), we show the
transformer output when the output weight values are set to zero except
the $(3,51)$ weight value. In other words, we check the output only
through the $(3,51)$ cell.  We can see that the output through this cell
is almost consistent with the least squares estimate, which implies that
the $(3,51)$ cell significantly contributes the entire output and, as in
our construction, the output of the $(3,51)$ cell 
approximates (\ref{eq:Q_m_3-N+1_3}) well. Note that, however, the output
through the $(3,51)$ cell is not entirely consistent with the
transformer output. This point will be mentioned in the conclusion.

\section{Conclusion and future work}

In this paper, we constructed the transformer that outputs the least
squares estimate of a simple regression problem. To achieve this, we
approximately represent the closed form solution by using the layer
normalization, which is an insight of \cite{Akyurek2022}.  For in-context
learning of regression tasks, most studies have considered
implementations of the gradient descent algorithms. This is because
multiplication, which is a basic operation of the attention mechanism,
is suitable for implementing the gradient descent algorithm.  However,
the experimental evidence that the gradient descent algorithm is
implemented through training is still unclear.  In contrast, in this
paper, we experimentally showed that there exist the examples, in which
our implementation based on the layer normalization is mainly used in
the transformer trained with the $\ell_1$ regularization when the target
output is the least sqaures estimate. Althogh this paper considered a
very simple case, it may be a first attempt to step into the training of
transformer that implements in-context learning. Therefore, we further have
several things to address.

\begin{itemize}
 \item Unfortunately, as see in Fig. \ref{fig:fitting_example}, the output
constructed based on the layer normalization is not entirely consistent
with the transformer output. Thus, the trained transformer has the other
mechanism to further fit the least squared estimate well. In
Fig. \ref{fig:weight_value} (b), we can see that the output weight
values that are assigned to the third embedding cells are relatively
larger than the weight values in the other embedding cells. Therefore,
the outputs of the third embedding cells may serve to complement the
approximation error in our construction. Also, as in Table
\ref{tbl:error-correlation}, the trained transformer in the fourth run
may not implement the division by $V_m$ using the layer normalization in
the second layer. Nevertheless, the test error is low. Furthermore, in
training without the $\ell_1$ regularization, similar results are
obtained. Thus, there may be different types of solver for obtaining
the least squares estimate for $\TF(4,2,2)$. Hence, we need
the investigation to reveal the other mechanism as part of our future
work.

\item Note that the gradient descent based implementation may not be
       comparable
in $\TF(4,2,2)$ since the number of layers of this transformer
may not be enough to implement the sufficient number of
gradient descent steps. For example, the two gradient descent steps can
be implemented by a $3$-layer transformer with linear attention; e.g.,
see \ref{sec:appendix-gd}.  The detailed comparison to the gradient
descent algorithm is left as a future work.

\item In our experiment, the prediction point is generated from the same
distribution of the training inputs. Therefore, the approximation under
a large $N$ assumption is almost accurate; i.e., for example, in
(\ref{eq:xi_1_1_2}), the mean of $N$ samples is almost consistent with
that of $N+1$ samples. Therefore, we need to investigate the case where
the prediction point is out-of-distribution. Note that this relates to
the generalization capability of transformers. On the other hand, as
noted in Sec. \ref{sec:discussion}, a large $N$ approximation may be
relaxed by the choice of the prompt format (design). The construction
under the other prompt format and the numerical investigation are parts
of our future work.

\item Although, in our experiment, we employed the least sqaures estimate as
the target output in training, the output sample at the prediction point
may be natural as the target output and it may be easily
       collectible. The experimental analysis of this case is also left
       as a future work.

\end{itemize}


\appendix

\section{An example of a lack of uniqueness}
\label{sec:appendix-nonuniqueness}

We show here a simple example, in which LSA produces the same output for
the different inputs.

Let $\Q$ be an $N\times 3$ matrix whose $n$-th row is $[1,a_n,b_n]$
and define $a:=\sum_{n=1}^Na_n$ and $b:=\sum_{n=1}^Nb_n$.
In (\ref{eq:LSA}), we set this $\Q$ as the input matrix and we set
$\vTheta=\{\W_1,\W_2,\W_3\}$, where
\begin{align}
\W_1=\W_3=\begin{bmatrix}
      1 & 0 & 0\\
      0 & 0 & 0\\
      0 & 0 & 0\\
     \end{bmatrix},~
\W_2=\begin{bmatrix}
      0 & 0 & 0\\
      0 & 1 & 0\\
      0 & 0 & 1\\
     \end{bmatrix}.
\end{align}
Then, the $n$-th row of $\LSA_{\vTheta}(\Q)$ is $[0,a,b]$ for any $n$.
On the other hand, let $\Q$ be an $N\times 3$ matrix whose $n$-th row is
$[1,a_n,a_n+b_n]$. In (\ref{eq:LSA}), by setting
\begin{align}
\W_1=\W_3=\begin{bmatrix}
      1 & 0 & 0\\
      0 & 0 & 0\\
      0 & 0 & 0\\
     \end{bmatrix},~
\W_2=\begin{bmatrix}
      0 & 0 & 0\\
      0 & 1 & -1\\
      0 & 0 & 1\\
     \end{bmatrix},
\end{align}
the $n$-th row of $\LSA_{\vTheta}(\Q)$ is $[0,a,b]$ for any $n$.
This is the same as in the first case.

\section{Implementation of gradient descent method}
\label{sec:appendix-gd}

\subsection{Gradient descent method for a simple regression}

The least squares estimates at $u_m$ is given by
(\ref{eq:lse-estimate-at-u_m}). Here, we consider to obtain this
solution by using a gradient descent method.  We define
\begin{align}
f_{\theta_m}(u_m):=\oy_m+\theta_m(u-\ox_m), 
\end{align}
where $\theta_m$ is a parameter to be adjusted. 
Thus, $\etheta_m=\vxy_m/\vxx_m$ is a closed form solution to $\theta_m$.
We also define 
\begin{align}
S_m(\theta):=\frac{1}{2}\sum_{n=1}^N(y_{m,n}-f_{\theta}(x_{m,n}))^2. 
\end{align}
We have
\begin{align}
&\frac{\partial S_m(\theta)}{\partial\theta}\notag\\
&=-\sum_{n=1}^N(y_{m,n}-f_{\theta}(x_{m,n}))(x_{m,n}-\ox_m)\notag\\
&=-\sum_{n=1}^N((y_{m,n}-\oy_m)-\theta(x_{m,n}-\ox_m))(x_{m,n}-\ox_m)\notag\\
&=-N(\vxy_m-\theta \vxx_m).
\end{align}
By setting $\theta_m(0)$ as an initial value, at the $t$-th step,  
the update equation of the gradient descent for $\theta_m(t)$ is given by
\begin{align}
\theta_m(t)&=\theta_m(t-1)-\eta
\left.\frac{\partial S_m(\theta)}{\partial\theta}\right|_{\theta=\theta_m(t-1)}\notag\\
&=\theta_m(t-1)+N\eta(\vxy_m-\theta_m(t-1)\vxx_m),
\end{align}
where $\eta>0$ is a learning rate.

We here give a solution to the second step. We set 
\begin{align}
\label{eq:gd-initial-value}
\theta_m(0)=a_m  
\end{align}
and have
\begin{align}
\theta_m(1)&=\theta_m(0)+N\eta(\vxy_m-\theta_m(0)\vxx_m)\notag\\
&=a_m+N\eta(\vxy_m-a_m\vxx_m)
\end{align}
By a simple calculation, we then have
\begin{align}
&\theta_m(2)\notag\\
&=\theta_m(1)+N\eta(\vxy_m-\theta_m(1)\vxx_m)\notag\\
&=a_m+2N\eta\vxy_m-2N\eta a_m\vxx_m-N^2\eta^2\vxy_m\vxx_m
+N^2\eta^2 a_m\vxx_m^2.
\end{align}
and we obtain the estimate at $u_m$ by
\begin{align}
\label{eq:2-step-gd-output}
f_{\theta_m(2)}(u_m)
=\oy_m+\theta_m(2)(u_m-\ox_m)
\end{align}
with this $\theta_m(2)$.

In this section, we show an implementation of this $2$-step gradient
descent by $\TF(4,3,2)$ under the assumption that $N$ is large or the
$O(1/N)$ term is negligible. Here, we omit the layer normalization to
simplify the construction and focus on the role of multiplication. We
here drop the subscript $m$ that denotes an index of training data for
the transformer.

\subsection{Input embedding}

We define
\begin{align}
\Win=\begin{bmatrix}
         1 & 0 & 0 \\
0 & \alpha_1 & 0\\
0 & \alpha_2 & \beta_2\\
0 & \alpha_3 & \beta_3\\
        \end{bmatrix}
\end{align}
and $\bin=\O_{D,1}$. By (\ref{eq:input-embedding}), we then have the
embedding whose $i$-th row is
\begin{align}
\Q_1[n,:]^\top=
\begin{bmatrix}
1\\ 
\alpha_1x_n\\
\alpha_2x_n+\beta_2y_n\\
\alpha_3x_n+\beta_3y_n\\
\end{bmatrix}.
\end{align}
$\Q_1$ is the $(N+1)\times D$ matrix and is the input to $\TF(4,3,2)$.

\subsubsection{The first layer}

We show a construction of the first head. 
Since $x_{N+1}=u$ and $y_{N+1}=0$, we have
\begin{align}
\Q_1[:,1]^\top \Q_1[:,3]
&=\sum_{n=1}^{N+1}1\cdot \left(\alpha_2x_n+\beta_2y_n\right)\notag\\
&=\alpha_2N\ox+\beta_2N\oy+\alpha_2u\notag\\
&\simeq\alpha_2N\ox+\beta_2N\oy\\
\Q_1[:,1]^\top \Q_1[:,4]
&=\sum_{n=1}^{N+1}1\cdot \left(\alpha_3x_n+\beta_3y_n\right)\notag\\
&=\alpha_3N\ox+\beta_3N\oy+\alpha_3u\notag\\
&\simeq\alpha_3N\ox+\beta_3N\oy
\end{align}
by omitting $O(1)$ terms under the assumption that $N$ is large.
If we set
\begin{align}
\W_{1,1,1}&=\begin{bmatrix}
         -\frac{1}{N} & 0 & 0 & 0 \\
0 & 0 & 0 & 0\\
0 & 0 & 0 & 0\\
0 & 0 & 0 & 0\\
        \end{bmatrix}\\
\W_{1,1,2}&=\begin{bmatrix}
         0 & 0 & 0 & 0 \\
0 & 0 & 0 & 0\\
0 & 0 & 1 & 0\\
0 & 0 & 0 & 1\\
        \end{bmatrix}\\
\W_{1,1,3}
&=\begin{bmatrix}
1 & 0 & 0 & 0\\
0 & 0 & 0 & 0\\
0 & 0 & 0 & 0\\
0 & 0 & 0 & 0\\
\end{bmatrix}
\end{align}
then, by (\ref{eq:LSA}), we have
\begin{align}
\H_{1,1}[n,:]^\top
=\begin{bmatrix}
0\\
0\\
-\alpha_2\ox-\beta_2\oy\\
-\alpha_3\ox-\beta_3\oy\\
\end{bmatrix}.
\end{align}

We next show a construction of the second head. We have
\begin{align}
\Q_1[:,1]^\top \Q_1[:,2]\simeq \alpha_1 N\ox
\end{align}
when $N$ is large. Thus, if we set
\begin{align}
\W_{1,2,1}&=\begin{bmatrix}
         -\frac{1}{N} & 0 & 0 & 0 \\
0 & 0 & 0 & 0\\
0 & 0 & 0 & 0\\
0 & 0 & 0 & 0\\
        \end{bmatrix}\\
\W_{1,2,2}&=\begin{bmatrix}
         0 & 0 & 0 & 0 \\
0 & 1 & 0 & 0\\
0 & 0 & 0 & 0\\
0 & 0 & 0 & 0\\
        \end{bmatrix}\\
\W_{1,2,3}
&=\begin{bmatrix}
1 & 0 & 0 & 0\\
0 & 0 & 0 & 0\\
0 & 0 & 0 & 0\\
0 & 0 & 0 & 0\\
\end{bmatrix}
\end{align}
then, by (\ref{eq:LSA}), we have
\begin{align}
\H_{1,2}[n,:]
=\begin{bmatrix}
0 & -\alpha_1\ox & 0 & 0\\\end{bmatrix}.
\end{align}

As a result, the output of the first layer is
\begin{align}
\Q_2=(\H_{1,1}+\H_{1,2})+\Q_1, 
\end{align}
whose $n$-th row is given by
\begin{align}
\Q_2[n,:]^\top=
\begin{bmatrix}
1\\
\alpha_1(x_n-\ox)\\
\alpha_2(x_n-\ox)+\beta_2(y_n-\oy)\\
\alpha_3(x_n-\ox)+\beta_3(y_n-\oy)\\
\end{bmatrix}
\end{align}
since we omit the layer normalization.

\subsection{The second layer}

We show a construction of the firs head. We have
\begin{align}
&\Q_2[:,1]^\top \Q_2[:,2]\notag\\
&=\sum_{n=1}^{N+1}
\alpha_1(x_n-\ox)\left\{\alpha_2(x_n-\ox)+\beta_2(y_n-\oy)\right\}\notag\\
&\simeq N\xi_{2,1}
\end{align}
for a large $N$, where
\begin{align}
\label{eq:gd-xi-_2_1}
\xi_{2,1}:=\alpha_1\alpha_2\vxx+\alpha_1\beta_2\vxy.
\end{align}
If we set
\begin{align}
\W_{2,1,1}&=\begin{bmatrix}
         0 & \frac{1}{N} & 0 & 0 \\
0 & 0 & 0 & 0\\
0 & 0 & 0 & 0\\
0 & 0 & 0 & 0\\
        \end{bmatrix}\\
\W_{2,1,2}&=\begin{bmatrix}
         0 & 0 & 0 & 0 \\
0 & 0 & 0 & 0\\
0 & 0 & 0 & 1\\
0 & 0 & 0 & 0\\
        \end{bmatrix}\\
\W_{2,1,3}&=\begin{bmatrix}
0 & 0 & 0 & 0\\
1 & 0 & 0 & 0\\
0 & 0 & 0 & 0\\
0 & 0 & 0 & 0\\
\end{bmatrix}
\end{align}
then, by (\ref{eq:LSA}), we have
\begin{align}
\H_{2,1}[n,:]
=\begin{bmatrix}
0 & 0 & 0 & \xi_{2,1}\alpha_1(x_n-\ox)\\
\end{bmatrix}.
\end{align}

We next show a construction of the second head.
If we set 
\begin{align}
\W_{2,2,1}=\W_{2,2,2}=\W_{2,2,3}=\O_{4,4},
\end{align}
then we formally have
\begin{align}
\H_{2,2}=\O_{N+1,4}.
\end{align}

Therefore, the the output of the second layer is
\begin{align}
\Q_3=(\H_{2,1}+\H_{2,2})+\Q_2, 
\end{align}
whose $n$-th row is given by
\begin{align}
\Q_3[n,:]^\top=
\begin{bmatrix}
1\\
\alpha_1(x_n-\ox)\\
\alpha_2(x_n-\ox)+\beta_2(y_n-\oy)\\
(\xi_{2,1}\alpha_1+\alpha_3)(x_n-\ox)+\beta_3(y_n-\oy)\\
\end{bmatrix}.
\end{align}

\subsection{The third layer}

We show a construction of the first head. We have
\begin{align}
&\Q_3[n,2]^\top \Q_3[n,4]\notag\\
&=\sum_{n=1}^{N+1}
\left\{(\xi_{2,1}\alpha_1+\alpha_3)(x_n-\ox)+\beta_3(y_n-\oy)\right\}
\alpha_1(x_n-\ox)\notag\\
&\simeq N\xi_{3,1}
\end{align}
for a large $N$, where
\begin{align}
\label{eq:gd-xi_3_1}
\xi_{3,1}&=
\alpha_1(\xi_{2,1}\alpha_1+\alpha_3)\vxx+\alpha_1\beta_3\vxy.
\end{align}

If we set
\begin{align}
\W_{3,1,1}&=\begin{bmatrix}
         0 & 1/N & 0 & 0 \\
0 & 0 & 0 & 0\\
0 & 0 & 0 & 0\\
0 & 0 & 0 & 0\\
        \end{bmatrix}\\
\W_{3,1,2}&=\begin{bmatrix}
         0 & 0 & 0 & 0 \\
0 & 0 & 0 & 0\\
0 & 0 & 0 & 0\\
0 & 0 & 1 & 0\\
        \end{bmatrix}\\
\W_{3,1,3}
&=\begin{bmatrix}
0 & 0 & 0 & 0\\
1 & 0 & 0 & 0\\
0 & 0 & 0 & 0\\
0 & 0 & 0 & 0\\
\end{bmatrix}
\end{align}
then we have
\begin{align}
\H_{3,1}[n,:]
=\begin{bmatrix}
0 & 0 & \xi_{3,1}\alpha_1(x_n-\ox) & 0\\
\end{bmatrix}.
\end{align}

We next show a construction of the second head. If we set 
\begin{align}
\W_{3,2,1}=\W_{3,2,2}=\W_{3,2,3}=\O_{4,4},
\end{align}
then we formally have
\begin{align}
\H_{3,2}=\O_{N+1,4}.
\end{align}
Thus, the the output of the third layer is 
\begin{align}
\Q_4=(\H_{3,1}+\H_{3,2})+\Q_3, 
\end{align}
whose $n$-th row is given by
\begin{align}
\Q_4[n,:]^\top=
\begin{bmatrix}
1\\
\alpha_1(x_n-\ox)\\
\xi_{3,1}\alpha_1(x_n-\ox)+\alpha_2(x_n-\ox)+\beta_2(y_n-\oy)\\
\xi_{2,1}\alpha_1(x_n-\ox)+\alpha_3(x_n-\ox)+\beta_3(y_n-\oy)\\
\end{bmatrix},
\end{align}
where we can write
\begin{align}
\xi_{3,1}&=
\alpha_1(\xi_{2,1}\alpha_1+\alpha_3)\vxx+\alpha_1\beta_3\vxy\notag\\
&=(\alpha_1^2\xi_{2,1}+\alpha_1\alpha_3)\vxx+\alpha_1\beta_3\vxy\notag\\
&=(\alpha_1^3\alpha_2\vxx+\alpha_1^3\beta_2\vxy
+\alpha_1\alpha_3)\vxx+\alpha_1\beta_3\vxy\notag\\
&=\alpha_1\alpha_3\vxx+\alpha_1\beta_3\vxy
+\alpha_1^3\alpha_2\vxx^2+\alpha_1^3\beta_2\vxy\vxx
\end{align}
by (\ref{eq:gd-xi-_2_1}) and (\ref{eq:gd-xi_3_1}).

$\Q_4$ is flattended into $\q$ which is a $4(N+1)\times 1$ vector.
Then, by setting
\begin{align}
\Wout[n]=
\begin{cases}
1 & n=4N+3\\
0 & {\rm otherwise},
\end{cases}
\end{align}
the prediction of the transformer is obtained by
\begin{align}
\label{eq:transformer-output-gd}
\ey(u)&=\Q_4[N+1,3]\notag\\
&=-\beta_2\oy+\xi_{4,1}(u-\ox),
\end{align}
where
\begin{align}
\xi_{4,1}
&=\alpha_2+\alpha_1^2\alpha_3\vxx+\alpha_1^2\beta_3\vxy
+\alpha_1^4\alpha_2\vxx^2+\alpha_1^4\beta_2\vxy\vxx.
\end{align}

\subsection{Correspondence to gradient descent}

By the correspondence between (\ref{eq:transformer-output-gd}) and
(\ref{eq:2-step-gd-output}), we have
\begin{align}
\begin{cases}
-\beta_2&=1\\
-\alpha_2&=a\\
\alpha_1^2\alpha_3&=-2n\eta a\\
\alpha_1^2\beta_3&=2n\eta\\
\alpha_1^4\alpha_2&=n^2\eta^2a\\
\alpha_1^4\beta_2&=-n^2\eta^2\\ 
\end{cases}
\end{align}
where $a$ is an initial value of the $2$-step gradient decent in (\ref{eq:gd-initial-value}).
By solving these equations, we have
\begin{align}
\label{eq:transformer-parameter-setting}
(\alpha_1,\alpha_2,\beta_2,\alpha_3,\beta_3)
=(\sqrt{n\eta},a,-1,-2a,2).
\end{align}
Therefore, the output of the transformer is approximately consistent with
the output of the linear function after the $2$-step gradient descent.


\end{document}